\documentclass[10pt,conference,a4paper]{IEEEtran}

\usepackage{graphicx}
\usepackage{amsmath,amssymb} 
\usepackage{color}
\usepackage{times}
\usepackage{epsfig}
\usepackage{graphicx}
\usepackage{url}
\usepackage{xcolor}
\usepackage{bm}

\newcommand\blfootnote[1]{%
  \begingroup
  \renewcommand\thefootnote{}\footnote{#1}%
  \addtocounter{footnote}{-1}%
  \endgroup
}

\newcommand{\specialcell}[2][c]{%
  \begin{tabular}[#1]{@{}c@{}}#2\end{tabular}}

\usepackage{hyperref}

\ifCLASSINFOpdf
\else
\fi
\begin{document}
%
\title{A Computational Approach to Relative Aesthetics}

\author{\IEEEauthorblockN{Vijetha Gattupalli$^*$, Parag S. Chandakkar$^*$ and Baoxin Li}
\IEEEauthorblockA{School of Computing, Informatics and Decision Systems Engineering, Arizona State University} \\
\tt\small \{jgattupa,pchandak,baoxin.li\}@asu.edu}



%


\IEEEoverridecommandlockouts
\IEEEpubid{\makebox[\columnwidth]{978-1-4799-7492-4/15/\$31.00~
\copyright2016
IEEE \hfill} \hspace{\columnsep}\makebox[\columnwidth]{ }}

\maketitle

\begin{abstract}
Computational visual aesthetics has recently become an active research area. 
Existing state-of-art methods formulate this as a binary classification task where a given image is predicted to be beautiful or not. In many applications such as image retrieval and enhancement, it is more important to rank images based on their aesthetic quality instead of binary-categorizing them. Furthermore, in such applications, it may be possible that all images belong to the same category. Hence determining the aesthetic ranking of the images is more appropriate. To this end, we formulate a novel problem of ranking images with respect to their aesthetic quality. We construct a new dataset of image pairs with relative labels by carefully selecting images from the popular AVA dataset. Unlike in aesthetics classification, there is no single threshold which would determine the ranking order of the images across our entire dataset. We propose a deep neural network based approach that is trained on image pairs by incorporating principles from relative learning. Results show that such relative training procedure allows our network to rank the images with a higher accuracy than a state-of-art network trained on the same set of images using binary labels.

\end{abstract}


%
\IEEEpeerreviewmaketitle

\section{Introduction} \label{sec:introduction}

Automatic assessment of image aesthetics is an active area of research due to its wide-spread applications. Most of the existing state-of-art methods treat this as a classification problem where an image is categorized as either beautiful (having high aestheticism) or non-beautiful (having low aestheticism)\footnote{We use this terminology throughout the paper.}\blfootnote{$^*$ indicates equal contribution by authors.}. In \cite{datta2006studying,ke2006design}, this problem has been formulated as a classification/regression problem by mapping an image to a rating value. Various approaches such as \cite{datta2006studying,ke2006design,bhattacharya2010framework,luo2008photo,dhar2011high,luo2011content,nishiyama2011aesthetic,o2011color,su2011scenic,marchesotti2011assessing}
have been proposed which either use photographic rules or hand-crafted features to assess the aesthetics of an image. Due to the recent success of deep convolutional networks, approaches such as \cite{lu2014rapid,lu2015deep} claim to have learned the feature representations necessary to categorize the given image as either beautiful or non-beautiful\blfootnote{Acknowledgement: The work was supported in part by ONR grant N00014-15-1-2344 and ARO grant W911NF1410371. Any opinions expressed in this material are those of the authors and do not necessarily reflect the views of ONR or ARO.}.

The approaches based on photographic rules have certain limitations. For example, the implementations of these rules may be an approximation, thus affecting the accuracy of aesthetic assessment. Also, the rules may not sufficiently govern the process of how we decide the aesthetic quality of an image. It is possible that some of the important rules have been left out or some erroneous ones have been included. These rules are mostly accompanied by generic image descriptors or task-specific hand-crafted features. Such approaches suffer from the disadvantages of generic/hand-crafted features that they may not be suited for a special task such as aesthetic assessment or the feature space does not fully represent the key characteristics which make an image aesthetic. The deep neural network based approaches may overcome these disadvantages by learning the feature representations.

While deep learning approaches have advanced the state-of-art for this task, we observe that classifying a given image as beautiful or non-beautiful may not always be the natural choice for some applications. It may also be more intuitive for humans to compare two images rather than giving an absolute rating to an image based on its aesthetic quality. Moreover, all images in a set could belong to the beautiful or non-beautiful category according to a classification model. In such cases, it may often be necessary to rank the images according to their aesthetic quality. For example, a machine-learned enhancement system \cite{yan2014learning} has to provide an enhanced version of the query image to the user. To do so, it needs to compare two images with respect to their aesthetics to determine which enhancement results in a more beautiful image. In an image retrieval engine, it would be desirable to have an option to retrieve images having low/similar/high aesthetic quality as compared to the query image. 

\IEEEpubidadjcol 

Motivated by these observations, we introduce a novel problem of picking a more beautiful image from a pair. We term this problem as ``Relative Aesthetics''. We build a new dataset of image pairs for this task by carefully choosing images from the popular AVA dataset \cite{murray2012ava} to satisfy certain constraints. For example, we observed that comparing images from unrelated categories (for example, a close-up of a car and a wedding scene) does not make sense and hence such pairs are avoided. There exists no single threshold which can binary-classify the pairs correctly across the entire dataset. In other words, if images were categorized into beautiful and non-beautiful, then some of the pairs in our data could contain both beautiful or both non-beautiful images. The details of dataset creation and its statistical analysis are provided in Section \ref{sec:dataset}.

Our problem draws certain parallels with ``relative attributes'' \cite{parikh2011relative}, where it was observed that training on relatively-labeled data leads to models that capture more general semantic relationships. They also mention that by using attributes as a semantic bridge, their model can relate to an unseen object category quite well. On the other hand, our problem presents different challenges. In \cite{parikh2011relative}, they compare two images with respect to attributes (for example, more natural, furrier, narrower etc.), which are better defined than the aesthetics of two images. Thus even though it is trivial to use models trained on categorical data to solve these ranking tasks, we found that using relative learning principles allows us outperform previous state-of-art classification models by gaining a more general and a semantic-level understanding of the proposed problem.

Our contributions are as follows:

\begin{enumerate}
\item We propose a novel problem termed as ``relative aesthetics'', which involves picking a more beautiful image from a given pair of images. We create a new dataset which has such relative labels from the popular AVA dataset by careful and constrained selection of image pairs. We will make our dataset and model source code publicly available upon the decision of the paper.

\item We build a deep network incorporating the relative learning paradigm and train it end-to-end. To the best of our knowledge, there is no prior work on studying aesthetics in a relative manner using deep neural networks.

\item We show that our model trained on relatively-labeled data is able to outperform a recent state-of-art method \cite{lu2014rapid} trained on a similar sized, categorically labeled dataset for the proposed task.

\end{enumerate}

Section \ref{sec:literature} discusses the relevant literature. Section \ref{sec:ourApproach} describes our relative, deep neural network based approach. Section \ref{sec:dataset} and \ref{sec:experiments} describe the data-creation, experimental setup, results and analysis. We conclude in Section \ref{sec:conclusion}.

\section{Related Work} \label{sec:literature}

Computational aesthetics research in the earlier years was focused on employing photographic rules, hand-crafted features or generic image descriptors. Intuitive and common properties such as color \cite{datta2006studying,nishiyama2011aesthetic,o2011color}, texture \cite{datta2006studying,ke2006design}, content \cite{luo2011content,dhar2011high}, combination of photographic rules, picture composition and hand-crafted features \cite{dhar2011high,luo2008photo,luo2011content} have been used. The most commonly used photographic rules include \textit{Rule of Thirds} used in  \cite{dhar2011high,luo2008photo,datta2006studying}. Other compositional rules include low depth of field, opposing colors etc. \cite{dhar2011high}. Common color features such as lightness, color harmony and distribution, colorfulness have been quantified for aesthetics assessment by computational models \cite{datta2006studying,nishiyama2011aesthetic,o2011color}. Texture features based on wavelets edge distribution, low depth of field, amount of blur have also been used \cite{ke2006design,dhar2011high}. Approaches specifically trying to model content in the image by detecting people \cite{luo2011content,dhar2011high,luo2008photo}, generic image descriptors such as SIFT \cite{lowe2004distinctive} have been proposed in \cite{dhar2011high}. Inspired by the then success of deep neural network on various tasks such as image classification \cite{krizhevsky2012imagenet,ciresan2012multi}, object segmentation \cite{Chen_2013_CVPR}, facial point detection \cite{Sun_2013_ICCV_Workshops}, Decaf features \cite{donahue2013decaf} for style classification \cite{karayev2014recognizing} etc., \cite{lu2014rapid} proposed a deep-learning-based approach to aesthetics assessment. This approach classifies a given image as beautiful or non-beautiful depending on the entire image as well as its local patches. Another such approach was presented in \cite{lu2015deep} where the authors aggregate the information from multiple patches in the multiple-instance-learning manner to improve the result of aesthetics assessment. Most of these approaches treat aesthetics assessment as a binary classification task, which may not always be the best choice for many applications, as discussed before.

The concept of training on relatively-labeled data to improve model performance and provide it with certain semantic understanding of the problem has been well-explored. The work on relative attributes \cite{parikh2011relative} predicts the relative strength of individual property in images. It allows for comparison with an unseen object category in the attribute space. Models learned in such a way enable richer text descriptions of images. Relative attribute feedback was used in conjunction with semantic language queries to improve the image search capability in \cite{kovashka2012whittlesearch}. There are many such applications where relative learning has explored a new dimension of the problem and improved the overall understanding of the model of a given task.

In this work, we propose to employ the relative learning principles for the task of image aesthetics assessment. This task is extremely subjective and have vaguely-defined properties than other general attributes like size, being more natural etc. To allow for learning using hand-crafted features, various datasets have been proposed such as \textit{Photo.net, DpChallenge.com, AVA} dataset. The first two datasets contain 20,278 and 16,509 images respectively\footnote{Datasets hosted on \url{ritendra.weebly.com/aesthetics-datasets.html}}, whereas the AVA dataset \cite{murray2012ava} contains 250,000 images. Thus we use AVA to form image pairs which in turn will facilitate the learning of our approach. We propose a Siamese deep neural network architecture \cite{bromley1993signature} with a relative ranking loss, which takes an image pair as input and ranks them with respect to their aesthetic quality. The back-propagation happens with the loss obtained from the ranking function, which, we believe, helps the network explore the attributes of certain images that make them more beautiful than others.

\section{Proposed Approach} \label{sec:ourApproach}

\begin{table*}
  \caption{Architecture of a column in our proposed network. Convolution is represented as (padding, \# filters, receptive field, stride)} \label{tab:columnArch}
  \centering
  \setlength{\tabcolsep}{0.53em}
	\begin{tabular}{c c c c c c c c c c c c}
	Padded Input           & Conv & Max-pooling & Conv & Max-pooling & Conv & Conv & Dropout & Dense & Dropout & Dense & Dropout \\[4pt]
	\hline
	\hline \\[-4pt]
	$3 \times 230 \times 230$ & $2, 64, 11, 2$ & $2 \times 2$ & $1, 64, 5, 1$ & $2 \times 2$ & $1, 64, 3, 1$ & $-, 64, 3, 1$ & $0.5$ & $1000$ & $0.5$ & $256$ & 0.5
	\\[4pt]
	\hline
	\end{tabular}
\end{table*}

\begin{figure*}
\centering
\includegraphics[width=0.75\textwidth]{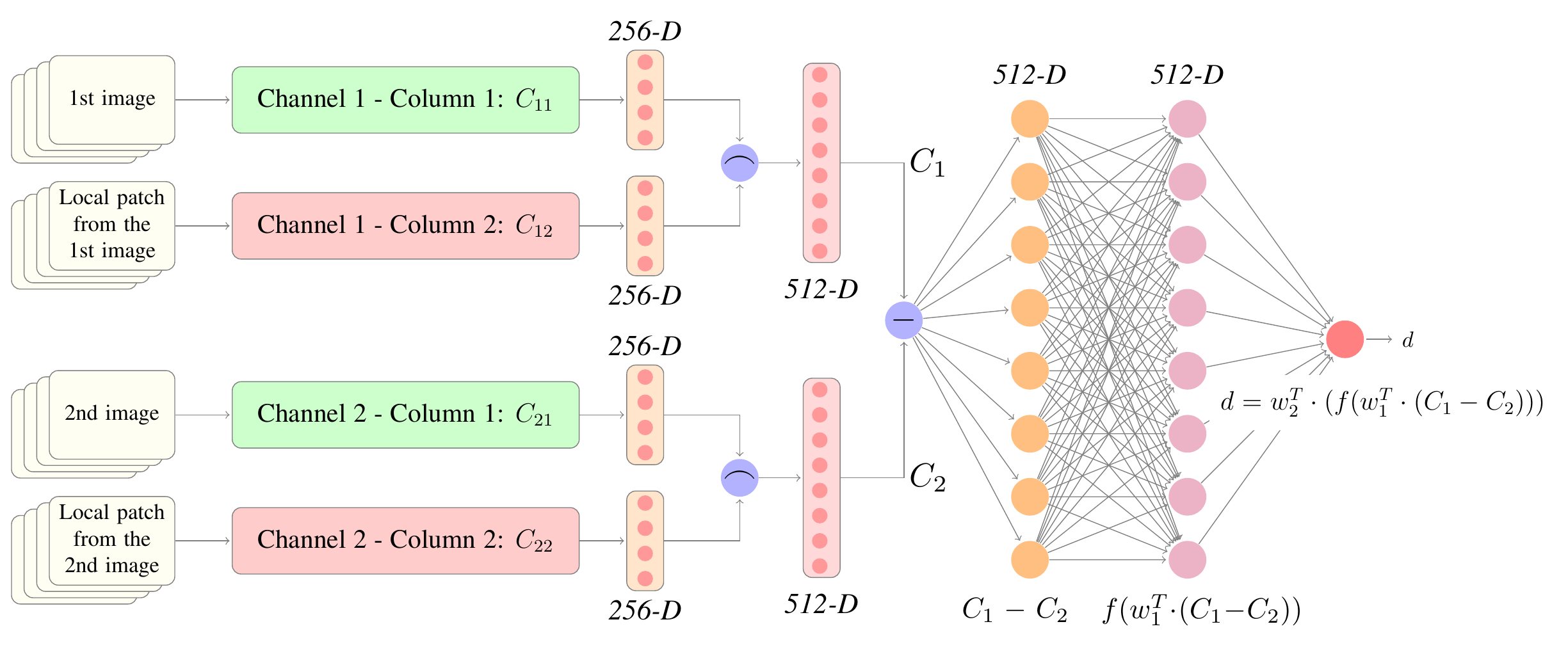}
\caption{Architecture of the proposed network; Weights are shared between the columns \(C_{11}\) and \(C_{21}\) (shown in green), \(C_{12}\) and \(C_{22}\) (shown in red); The features obtained from \(C_{11}\) and \(C_{12}\) are concatenated (represented by \(\frown\) symbol) to get \(C_1\) and \(C_{21}\) and \(C_{22}\) are concatenated to get \(C_2\); The vector \(C_1 - C_2\) is passed through two dense layers to obtain a score \(d\) comparing the aesthetics of two images. \(f(\cdot)\) denotes an ReLU non-linearity. Please refer to the text for further details.}
\label{fig:arch}
\centering
\end{figure*}

The comparison of the aesthetics of two images is dependent on many factors and people's visual preferences. Some of the factors include color harmony \cite{nishiyama2011aesthetic}, colorfulness \cite{datta2006studying}, inclusion of opposing colors \cite{dhar2011high}, composition \cite{litzel1975photographic}, visual balance \cite{niekamp1981exploratory} etc. They are also affected by the content in the picture \cite{luo2008photo,luo2011content}. Though determination of aesthetics is a subjective process, there are some well-established rules in the photography community such as low depth-of-field, rule of thirds, golden ratio \cite{joshi2011aesthetics}. However, making hand-crafted features for such rules is difficult and often will lead to approximation or misrepresentation of those rules. Therefore, we take a deep neural network based approach in which we incorporate relative ranking by designing a suitable loss function. Most of the rules or aesthetic criteria can be defined using either an entire image or a part of it. Therefore, for each image in the pair, our network is trained on two \textit{views} of an image as also done in \cite{lu2014rapid}: the entire image and a local patch. This enables the network to see different aspects of the input. For example, a view of the entire image may provide the network with the knowledge of color composition while the local patch may help with resolution, depth-of-field etc. We now describe our network architecture and its training procedure in detail.

\subsection{Network Architecture} \label{subsec:networkArch}

Our deep convolutional neural network (DCNN) takes an image pair as input. For each image in the pair, it takes as input that image itself and its local patch. Since all images have to be of the same size, they are warped to be $224 \times 224 \times 3$. A same-sized local patch is cropped from the original image. We choose to warp the image based on the findings in \cite{lu2014rapid}, which shows that local patches along with warped image gives the best result. Our network has two ``channels'' as shown in Fig. \ref{fig:arch}, corresponding to the input pair of images. A channel is defined as the part of our CNN which takes an image along with its local patch as input. Each channel has two ``columns''. One column takes the warped image and the other one takes its local patch as input.

Our architecture is a Siamese network where each channel shares weights in a certain way, which is shown in Fig. \ref{fig:arch} by means of color coding. The columns with the same color (i.e. either red or green) share the same weights. This is because the ranking produced by the network should be invariant to the order of the images in the pair. Both channels have exactly the same architecture until they are merged at the final dense layer of $512-D$. We now describe the architecture of the upper channel (channel 1). This channel has two columns which takes the image and its local patch as input. Since these two inputs are on a different spatial scale and trying to convey different aesthetic properties as discussed earlier, we do not set constraints on the weights of both the columns in a channel. The upper column in channel 1 ($C_{11}$) takes the entire image as input which is of size $224 \times 224 \times 3$, zero-padded with 3 pixels on all sides. The column has five convolutional layers. The first convolutional layer has 64 filters each of size $11 \times 11 \times 3$ with stride $2$. Second convolutional layer has 64 filters of size  $5 \times 5$ with stride 1. Third and fourth layer have 64 filters of size $3 \times 3$ with stride 1. These are followed by two dense layers of size 1000 and 256 respectively. We apply $50\%$ Dropout at these two dense layers. Max-pooling is applied after first two convolutional layers. Each max-pooling operation halves the input in both the directions. We use ReLU activation throughout. The architecture of $C_{11}$ is also detailed in Table \ref{tab:columnArch}. The lower column of channel 1 ($C_{12}$) and both the columns of channel 2 (i.e. $C_{21}$ and $C_{22}$) have the same architecture as $C_{11}$ including dropout, max-pooling and zero-padding operations. 

The key thing to note here is that the weights are shared for (i) the two columns which take the entire image as input i.e. $C_{11}$ and $C_{21}$, and (ii) the remaining two columns which take the local patches as input i.e. $C_{12}$ and $C_{22}$. $C_{11}$ and $C_{21}$ each generate a $256-D$ representation (i.e. of the entire image). Similarly, $C_{12}$ and $C_{22}$ also generate $256-D$ features (i.e. of the local patch). We concatenate the two $256-D$ representations from $(C_{11}, C_{12})$ as well as $(C_{21},C_{22})$ to form two $512-D$ representations. Fig. \ref{fig:arch} shows this architecture and the sharing of weights. 

Now we explain our ranking loss function which takes the above two $512-D$ representations and gives a quantitative measure comparing the aesthetics of the two images in a pair.

\subsection{Ranking Loss Layer} \label{subsec:rankingLossLayer}

Our network aims at correctly ranking two input images based on their underlying aesthetic quality. Formally, given two input images $I_1$ and $I_2$, we  decide that $I_1$ is more beautiful than $I_2$ (also denoted as $I_1 > I_2$ here onward) if a positive value is obtained for $d(I_1,I_2)$ and vice versa. In other words, $d(I_1,I_2)$ is a measure comparing aesthetics of two images.

\begin{equation} \label{eq:rankingLoss}
d(I_1, I_2) = w^T\cdot(g(I_1) - g(I_2))
\end{equation}

\noindent Here, $g(I_1)$ and $g(I_2)$ are the CNN representations. In our network, $g(I_1)$ and $g(I_2)$ are represented by $C_1$ and $C_2$ respectively, as shown in Fig. \ref{fig:arch}. To increase the representational power, we pass $(C_1-C_2)$ through two dense layers separated by a ReLU non-linearity. Thus for our network, Equation \ref{eq:rankingLoss} takes a slightly modified form as follows:

\begin{equation} \label{eq:rankingLoss_mod}
d(I_1, I_2) = w_2^T\cdot(f(w_1^T\cdot(C_1 - C_2)),
\end{equation}
\noindent where $f(\cdot)$ denotes an ReLU non-linearity.

Keeping this in mind, we can now design our final loss function with the following properties:

\begin{enumerate}
\item It should propagate zero loss when all image pairs are ranked ``correctly'' (i.e. the representations of the images in these pairs are separated by a margin $\delta$).
\item It should only be able to produce a non-negative loss.
\end{enumerate}

\noindent Hence the loss function is designed as follows:

\begin{equation}
L = \max(0,\delta-y \cdot d(I_1,I_2)),
\end{equation}

\noindent where $y$ is a ground-truth label which takes value $1$ if the first image in the pair is more beautiful than the second (i.e. $I_1 > I_2$) and it equals -1 if $I_1 < I_2$. The term $\max(0,\cdot)$ is necessary to ensure that only non-negative loss gets back-propagated. The $\delta$ is a user-defined parameter which serves two purposes. First, it defines a required separation to declare $I_1 > I_2$ (or $I_1 < I_2$). That means if $y \cdot d(I_1,I_2) > \delta$, then no loss should be back-propagated for such pairs. Secondly, and more importantly, $\delta > 0$ avoids a trivial solution to our optimization objective. To clarify further, if $\delta=0$, then for $y=1$ and $y=-1$, a common trivial solution exists which makes either $w_1=0$ or $w_2=0$. We set $\delta = 3$ as we do not find any performance boost by further increasing the separation between CNN feature representations of $I_1$ and $I_2$.

In the further subsections, we explain the training and testing procedures of our network. Then we compare the aesthetic ranking results of our network against a state-of-art network that is trained on a categorical data.

\subsection{Training Our Architecture} \label{subsec:training}

This architecture is trained using mini-batch SGD with a learning rate of $0.001$, momentum = 0.9, weight decay of $10^{-6}$ and by employing Nesterov momentum. The learning rate is reduced by $15\%$ after every $10$ epochs. The batch size is set to 50. Apart from warping and cropping out the local patch, we only subtract the mean RGB value computed on the training set from each pixel of the image. During training, when the network makes a wrong decision, it is forced to learn by exploiting the difference between some other characteristics of the image in the next iteration. Over a number of epochs, it manages to discover the relevant image properties which better define image aesthetics.

We have $23,000$ image pairs containing all unique images (i.e. total $46,000$ images). We use subsets of $20,000$ and $3,000$ pairs for training and validation respectively. We stop our training when the accuracy on the validation set does not show significant improvement for $10$ consecutive epochs. We train using relative labels i.e. a pair is labeled as 1 if $r_1 - r_2 > 1$, otherwise it is labeled as $-1$. Here, $r_i$ is the average rating of $I_i$ in AVA dataset. More details on the data creation are given in Section \ref{sec:dataset}.

\subsection{Testing Our Architecture} \label{subsec:ourTesting}

Given a new pair of images, we first subtract the mean of the training data from each pixel of both the images. We would like to point out that our test set does not share any pairs or any individual images with the training and validation set. We first pass both the images and their patches into our network and get the value of $d(I_1,I_2)$ from Equation \ref{eq:rankingLoss_mod}. $I_1$ is then predicted as a more beautiful image than $I_2$ if $d(I_1,I_2) > 0$ and vice versa. Our test set contains $20,000$ image pairs. We use the weights of the epoch where we achieve highest ranking accuracy with the least amount of loss on the validation set.

\subsection{Ranking using a Network Trained on Categorical Labels} \label{subsec:rankingUsingCatLabels}

We train a network on categorically-labeled data using our own implementation of the RAPID approach \cite{lu2014rapid}, which is a recent state-of-art method for aesthetics assessment. It is trained on the same set of 40,000 images that is used to train our network. However, in this case, these images have been categorized as either beautiful or non-beautiful depending on the average ratings obtained directly from the AVA dataset. We set the threshold that determines the class of an image equal to 5.5, since the ratings in the AVA dataset range from 1-10. This network consists of stacks of convolutional layers, followed by dense layers and finally a sigmoid to convert the raw scores into a probability measure, $p(y=1|I)$, i.e. probability of an image $I$  belonging to the beautiful class. We point the reader to \cite{lu2014rapid} for more details about the RAPID network architecture. While testing for a pair of input images, we pass first image through the network and get the probability measure - $p(y=1|I_1)$. Passing the second image gives us $p(y=1|I_2)$. We decide that first image is more beautiful than the second one if $p(y=1|I_1) > p(y=1|I_2)$. This test set contains $20,000$ image pairs  and is identical to the test set used for our approach as mentioned in Section \ref{subsec:ourTesting}. Despite RAPID network being similar in size to our network, it gets a significantly lower accuracy on this relative ranking problem, which suggests that a network trained on categorically-labeled data fails to learn the complex, relative ranking order in the data.


\section{Dataset} \label{sec:dataset}

Our task is to determine the more beautiful image in a pair. To the best of our knowledge, there exists no such dataset containing relatively-labeled pairs with respect to their aesthetic rating. We created a dataset containing $43,000$ image pairs. The individual images in these pairs belong to the AVA dataset \cite{murray2012ava}. We use $20,000$ pairs for training, $3,000$ for validation and the rest for testing. We now describe the protocol used to form the pairs out of the images from the AVA dataset. The protocol can be defined by these three constraints:

\noindent 1. The difference between the average ratings of images in a pair should be $\geq 1$. Constraining this difference ensures that the training/test pairs are more likely to be aesthetically different. 

\noindent 2. Each image in the AVA dataset has 210 ratings on an average. We computed variance of all the ratings for each image. We observed that the distribution of all these variances over the entire the AVA dataset takes the form of a Gaussian with a mean of 2.08 and a standard deviation of 0.6. The minimum and maximum variance in the image ratings are 0.8 and 4.5 respectively. As mentioned in \cite{murray2012ava}, high variances among the image ratings are a result of the collective disagreement between the raters, which suggests that such images may have certain abstract/novel content or photographic style, preferred only by certain group of people. We avoid the images which cause such significant disagreements among the raters by only considering the images having rating-variance less than 2.6.

\noindent 3. We avoid including pairs from different categories since the characteristics which make an image aesthetic may vary with the category. For example, a beautiful picture of a car may have bright colors whereas a beautiful picture of a human face may have low-depth of field and better details. Additionally, since the ratings in the AVA dataset are crowd-sourced ratings, the opinions may exhibit a preference towards some category. We can mitigate the effect of these two factors by using pictures from the same category to form pairs.

After such selection of pairs, we can form the relative labels. We label a pair as $1$ if the average rating of the first image is greater than that of the second image and $-1$ otherwise. The majority of the pairs in our dataset have the rating-difference $\approx$ 1. To quantify, the rating-difference for about $85\%$ of the training and test data is between 1 and 1.5. As the rating difference between the images of a pair decreases, choosing the more beautiful image in that pair gets difficult. Also, to ensure that our network is not biased towards our dataset, we replicate our experiments on another reference test-set provided by the creators of the AVA dataset \cite{murray2012ava}. This reference test-set contains $20,000$ images and has also been used by \cite{lu2014rapid}. By following the aforementioned protocol, these $20,000$ images yield us $7,670$ pairs. We call these set of pairs as the standard test set. We now describe the experiments and give analysis of results.

\section{Experiments and Results} \label{sec:experiments}
\begin{figure*}[!t]
\centering
\includegraphics[width = 0.78\textwidth]{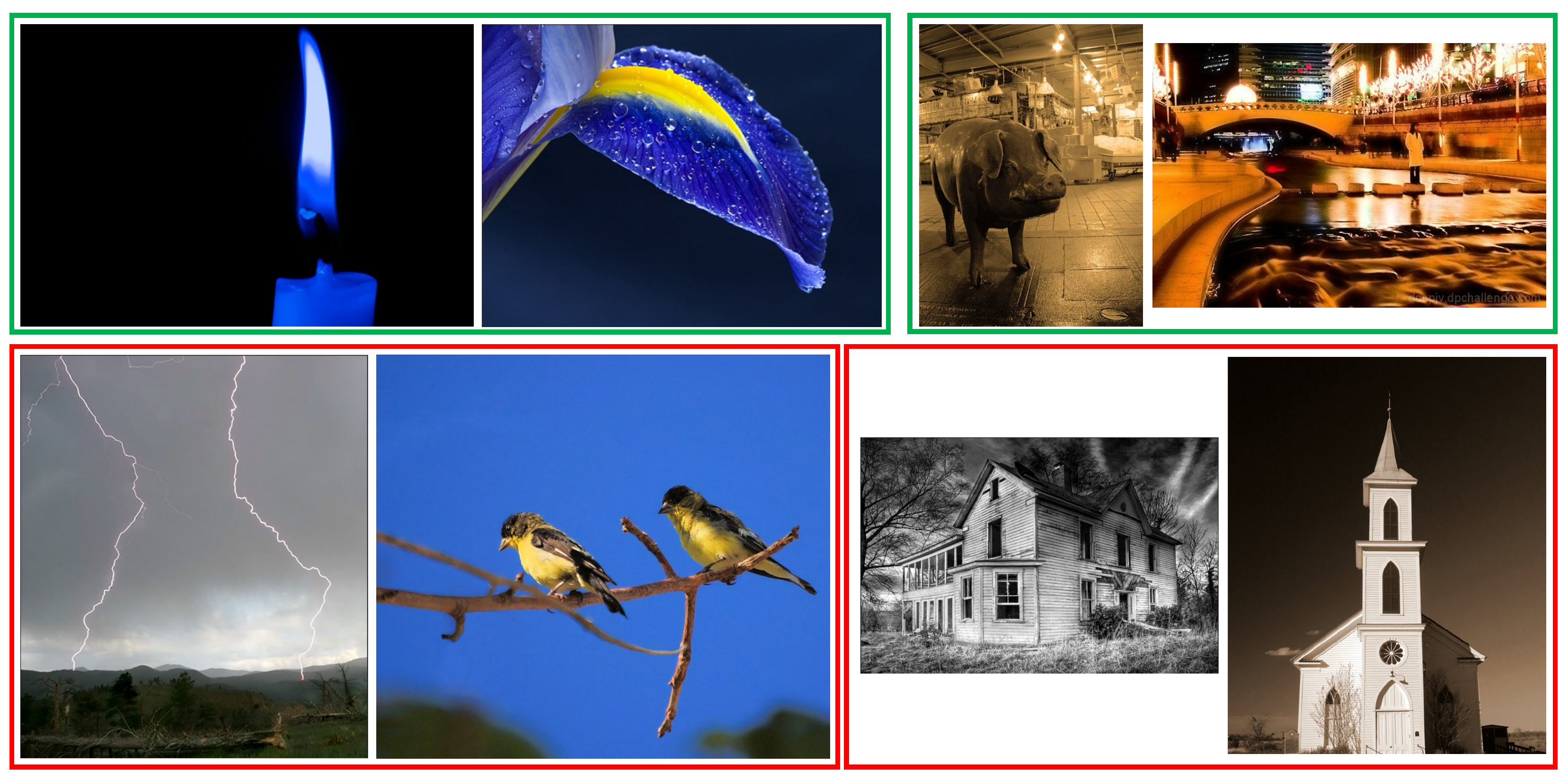}
\caption{Rankings produced by our network are shown above. Top and bottom rows show correct and wrong predictions respectively for a total of 4 pairs. Each of them are enclosed in either red/green boxes. For every pair, our network ranks the right image higher than the left image. Please view in color.}
\label{fig:predictions}
\centering
\end{figure*}
\noindent
\begin{table}[!t]
\vspace{4pt}
\caption{Results for ranking and binary classification}
\label{tab:results}
\setlength{\tabcolsep}{0.515em}
\begin{tabular}{c  c  c  c  c}
\hline \hline \\[-6pt]

 & \specialcell{Ranking \\ on our \\ test-set} &  \specialcell{Ranking on \\ the pairs from \\ standard test-set} & \specialcell{Classification \\ on our test-set} & \specialcell{Classification \\ on standard \\ test-set}\\
 \hline \\[-6pt]

RAPID \cite{lu2014rapid} & 62.21 & 65.87 & \textbf{59.92} & 69.18 \\[1pt]

Proposed & \textbf{70.51} & \textbf{76.77} & 59.41 & \textbf{71.60} \\[1pt]
\hline \hline

\end{tabular}
\end{table}

We run our network on our test set and the standard test set containing $20,000$ and $7,670$ image pairs respectively. We achieve a ranking accuracy of $70.51\%$ and $76.77\%$ on our test-set and on the standard test-set respectively. Here, ranking accuracy is defined as the fraction of pairs for which the model correctly picks the more beautiful image according to the ground-truth labels. We compare our approach with a state-of-art aesthetics classification network called RAPID \cite{lu2014rapid}, trained as described in Section \ref{subsec:rankingUsingCatLabels}: we pass both the images one-by-one to the RAPID network and choose the more beautiful image. RAPID produces a ranking accuracy of $62.21\%$ and $65.87\%$ on ours and the standard test-set respectively. Since each channel of our architecture is a replica of \cite{lu2014rapid} with the modified ranking loss, we compare our architecture only with \cite{lu2014rapid}. However, we believe that we will obtain similar performance improvements if a different state-of-art model (e.g. \cite{lu2015deep}) was used for each of our channels. 

Due to our relative-learning-based approach, we believe that our network has gained a semantic-level understanding of the properties which make an image highly aesthetic. To verify this, we attempted binary classification on our dataset as well as the standard test-set. For this purpose, we extracted the top channel of our network i.e. $C_{11}$ and $C_{12}$ (see Fig. \ref{fig:arch}). We use the best weights learned from the ranking task for this channel. After the last node, we just append a sigmoid layer to convert the values into decision values. The input image is passed through the network to obtain the probability of that image being beautiful. We compute our results on a subset of $10,000$ images taken from our test set and the entire standard test set \cite{murray2012ava} containing $20,000$ images. On our test set, proposed approach obtains $59.41\%$ classification accuracy as compared to $59.92\%$ obtained by RAPID. On the standard test set, we obtain an accuracy of $71.60\%$ as compared to $69.18\%$ obtained by RAPID. \textit{Note that we do not perform any training to adopt our network for classification}, which shows that the learned features may be capturing the characteristics that are responsible for making an image aesthetic. Our network outperforms RAPID on the ranking task and produces competitive performance on the classification task without any additional training. We note that the performance of both the networks is significantly lower on our test-set as compared to that of on the standard test-set. This performance difference could be attributed to the fact that all images in the standard test-set are distributed only over 8 categories, whereas the images in our test-set are distributed over all 65 categories. The results of all the experiments are summarized in Table \ref{tab:results}

Fig. \ref{fig:predictions} illustrates some ranking results obtained by our network. The wrong predictions in the bottom row show that the network lacks semantic knowledge about objects and natural phenomena. For example, even though the picture containing two birds has better color harmony/contrast, the lightning phenomena is a rare capture, making it more picturesque.

\section{Conclusion} \label{sec:conclusion}
We introduced a novel problem of relative aesthetics which could have widespread applications in image search, enhancement, retrieval etc. We created a dataset with a careful and constrained selection of $43,000$ pairs of images from the AVA dataset where one image is always more beautiful than the other. We showed that a deep neural network trained with an appropriate loss function which accounts for such relatively-labeled data, significantly outperforms a state-of-art network trained on same data with categorical labels. The proposed network is also able to achieve a competitive performance on an aesthetics classification problem with trivial modifications to its architecture and no fine-tuning at all. This shows that it has gained a certain semantic-level understanding of the factors involved in making an image aesthetic.

\bibliographystyle{IEEEtran}
\bibliography{egbib}
\end{document}